\documentclass{article}

\usepackage[nonatbib,final]{neurips_2018}

\usepackage[utf8]{inputenc} 
\usepackage[T1]{fontenc}    
\usepackage[colorlinks=true,linkcolor=blue,citecolor=blue]{hyperref}       
\usepackage{url}            
\usepackage{booktabs}       
\usepackage{amsfonts}       
\usepackage{nicefrac}       
\usepackage{microtype}      
\usepackage{graphicx}
\usepackage{helvet}
\usepackage{courier}
\usepackage{amsmath}
\usepackage{amssymb}
\usepackage{tabularx}
\usepackage{varwidth}
\usepackage{xcolor}
\usepackage{graphicx}
\usepackage{bbm}
\usepackage{bm}
\usepackage{colortbl}
\usepackage[clock]{ifsym}
\usepackage{enumitem}
\usepackage{multicol} 
\usepackage[utf8]{inputenc}
\usepackage{booktabs}  
\usepackage{algorithm}
\usepackage{algorithmic}
\usepackage{wrapfig}
\usepackage{amsthm}
\usepackage{multirow}
\usepackage{subfig}
\usepackage{setspace}
\usepackage{pifont}
\usepackage[export]{adjustbox}

\newcolumntype{C}[1]{>{\centering\arraybackslash}m{#1}}
\newcolumntype{R}[1]{>{\raggedright\arraybackslash}m{#1}}

\usepackage[font=small,skip=10pt]{caption}
\usepackage{setspace}
\DeclareCaptionFormat{myformat}{#1#2#3\hrulefill}

\title{Policy Analysis using Synthetic Controls in Continuous-Time}

\author{
  Alexis Bellot$^{1,2}$\hspace{0.5cm} Mihaela van der Schaar$^{1,2,3}$\\
  $^{1}$University of Cambridge, $^{2}$The Alan Turing Institute, $^{3}$University of California Los Angeles\\
  \texttt{[abellot,mschaar]@turing.ac.uk} \\
}

\begin{document}

\maketitle

\begin{abstract}
Counterfactual estimation using synthetic controls is one of the most successful recent methodological developments in causal inference. Despite its popularity, the current description only considers time series aligned across units and synthetic controls expressed as linear combinations of observed control units. We propose a continuous-time alternative that models the latent counterfactual path explicitly using the formalism of controlled differential equations. This model is directly applicable to the general setting of irregularly-aligned multivariate time series and may be optimized in rich function spaces -- thereby substantially improving on some limitations of existing approaches.
\end{abstract}

\section{Introduction}
Counterfactual estimation poses the question of what would have been the outcome if a different intervention had been applied. To answer this question one often seeks a control group of \textit{comparable} units e.g., individuals, patients, or states, to approximate a target unit's outcome trajectory had a different treatment been applied. 

We focus on the case where a \textit{single} target unit adopts the treatment or policy of interest at a particular point in time, and then remains exposed to this treatment at all times afterwards. Both pre-treatment and post-treatment outcomes are assumed to be available. We ask whether we can infer the counterfactual trajectory over time had the unit not been exposed to the treatment using a population of control units never exposed to the treatment.

The synthetic control method \cite{abadie2010synthetic,abadie2003economic} is one of the most important recent innovations in causal inference to solve this problem. It recognizes that a weighted combination of control units (instead of the standard procedure of seeking a single control or average in a neighbourhood of controls) often provides a more informative comparison for treatment effect estimation and then formalizes the selection of the comparison units using a data driven procedure. 

Synthetic controls have become widely popular in the fields of policy analysis due to their simplicity and transparency. They have been used to assess the effect of tax hikes on the consumption of cigarettes \cite{abadie2010synthetic,abadie2019using}, of drug programs on drug use and crime \cite{robbins2017framework}, of immigration policies \cite{borjas2017wage,bohn2014did}, of minimum wages \cite{allegretto2017credible}, of terrorism on economic growth \cite{abadie2003economic}, of large political changes and events \cite{hope2016estimating}, and also frequently in the biomedical sciences for estimating public health interventions \cite{pieters2016effect,bouttell2018synthetic}.

\subsection{The synthetic control method}
The typical setting considers $n$ units $Y_i = (Y_{i,t_1}, \dots, Y_{i,t_m}) \in \mathbb R^m$, $i=1, \dots, n$, each observed on the same grid of time points $t_1,\dots,t_m$. By convention, and without loss of generality, one unit ($i=1$) receives the treatment or intervention at time $T\in (t_0,t_m)$ while the rest act as the control group. Let $Y_{i,t}^0$ be the potential outcome for unit $i$ at time $t$ in a hypothetical world where the intervention did not occur, and analogously let $Y^1_{i,t}$ be the corresponding potential outcome assuming the intervention did occur. Both are functions evaluated at time $t$. 

The observed outcome for unit $i$ at time $t$, denoted by $Y_{i,t}$, therefore satisfies:
\begin{align}
\label{potential_outcomes}
    Y_{1,t} &= Y^0_{1,t} + (Y^1_{1,t} - Y^0_{1,t})D_{1,t},\nonumber\\
    Y_{i,t} &= Y^0_{i,t}, \qquad i=2,\dots,n
\end{align}
where $D_{1,t}$ is a binary indicator of whether unit $1$ is treated at time $t$, taken to be 0 at all times before treatment at time $T$ and 1 at all time after time $T$. $\tau_{1}(t) = Y^1_{1}(t) - Y^0_{1}(t)$ is the causal effect of intervention on unit $1$ at time $t$. 

Synthetic control methods suppose that there exist weights $w_2,\dots, w_{n}$ such that $Y^0_{1,t}$ can be written as a weighted average of observed control outcomes: $Y^0_{1,t} =\sum^{n}_{i=2} w_i Y_{i,t}$, for $t \in [t_0, T)$ before treatment assignment. And then use this approximation to compute the causal effect of the intervention:
\begin{align}
\label{te_eq}
    \hat\tau_{1}(t) = Y_{1,t} - \sum^{n}_{i=2} w_i Y_{i,t},
\end{align}
for every $t > T$. 

The time series defined by $\sum^{n}_{i=2} w_i Y_{i,t}$ is the \textbf{synthetic control}. 

It is called synthetic control because it is constructed such as to be representative of the treated unit ($i = 1$) had the treated unit not received treatment, and can be justified by assuming an underlying linear data generating mechanism for the data (with or without observed or unobserved confounders) \cite{abadie2010synthetic}.

A large and varied set of strategies for the estimation of $(w_2,\dots,w_n)$ have been proposed. We review some of these in section \ref{sec_related_work}.

\subsection{Limitations from a dynamical systems perspective}
We seek to improve upon two limitations of discrete-time synthetic controls.

\begin{enumerate}[leftmargin=*]
    \item In reality, the time series $(Y_{i,t_1}, \dots, Y_{i,t_m})$ is often assumed to be sequence of observations from an underlying \textit{continuous} process. Synthetic controls may then be interpreted as a discrete approximation of the latent counterfactual path of the treated unit. However, this approximation typically breaks down if units are misaligned in time or irregularly sampled. An issue that may be solved only imperfectly by discarding information or interpolating the data.
    
    \item Moreover, for complex problems, the \textit{linearity} of control combinations may be restrictive. And, for general dynamical systems the assumption that the outcome correlations $(w_2,\dots,w_n)$ are \textit{static and invariant over time} is not plausible. For instance, the correlation between control and treated units may change over time if driven by weakly coupled dynamical systems. Examples of such systems arise in social sciences \cite{ranganathan2014bayesian} and biology \cite{heltberg2019chaotic}. For extrapolation over time to be consistent these dynamics should be captured.
\end{enumerate}


\subsection{Contributions}
In this paper we take a very different approach to synthetic control estimation rooted in the theory of dynamical systems.

We propose to model the synthetic control as the solution to a \textit{controlled differential equation} \cite{lyons2007differential},
\begin{align}
\label{cde_contributions}
    dY^0_{1,t} = f \left(dY_{2,t},\dots,dY_{n,t},Y^0_{1,t}\right), \quad Y^0_{1,t_0}=y_{1,t_0},
\end{align}
where $y_{1,t_0}$ is an initial value and $f$ is a latent vector field that is learnt from data and serves to combine control paths to approximate the counterfactual dynamics of the treated unit. 

By integrating both sides, we construct a \textbf{continuous-time synthetic control} that is driven by a combination of the latent paths of control units. We thus retain the interpretation of equation (\ref{te_eq}) with the exception that we model a combination of the \textit{latent infinitesimal variation} of control outcomes instead of the explicit \textit{discrete-time observations} of control outcomes. 

This model has three key features. 
\begin{enumerate}[leftmargin=*, itemsep=0pt]
    \item It is capable of processing irregularly aligned data and may be evaluated at any point over time.
    \item $f$ may be modelled in rich function spaces that may capture non-linearities and varying dependencies between units over time.
    \item It may be trained efficiently with existing adjoint backpropagation algorithms and is easy to implement -- thereby offering a practical, fully non-parametric, and continuous-time alternative to existing synthetic control methods.
\end{enumerate}

\begin{figure*}[t]
\vspace*{0.5cm}
\centering
\includegraphics[width=1\textwidth]{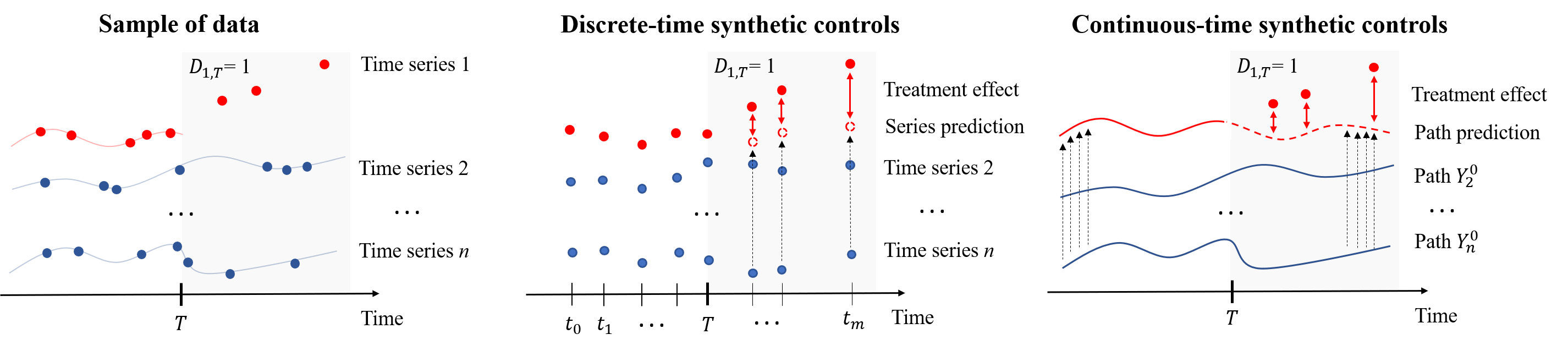}
\caption{\textbf{Left panel}: Some data process is observed at misaligned observation times. The problem is to approximate the counterfactual trajectory of Time series 1 after time $T$. \textbf{Middle panel}: Previous work typically requires aligned observation times and synthetic controls $\sum^{n}_{i=2} w_i Y_{i,t}$ are defined to depend discretely and linearly on control observations. \textbf{Right panel}: In contrast, the proposed continuous-time synthetic control $Y_{1,T} + \int_{T}^{t} f(Y^0_{1,s})\hspace{0.1cm} d\mathbf Y^0_s$ depends continuously on control paths over time and naturally accomodates for misaligned observations and complex dynamics.}
\label{overview}
\end{figure*}

\section{Problem formulation}
This section extends the formulation of synthetic controls to a more general time series setting, illustrated in Figure \ref{overview}. 

Suppose that each latent path $Y_{i}:[t_0,t_m] \rightarrow \mathbb R^d$ is partially-observed through $m$ irregular time series samples, $\left\{(t_{0}, Y_{i,t_0}),(t_{1}, Y_{i,t_1}), \dots, (t_{m}, Y_{i,t_m})\right\}$, with each $t_{j} \in \mathbb R$ the timestamp of the observation $Y_{i,t_j} \in \mathbb R^d$, and $t_{0} < \dots < t_{m}$. To avoid notation clutter, the time subscript refers to function evaluation and the case where each $i$-th observation sequence has its own $m_i$ irregular time stamps $t_{i,0},\dots, t_{i,m_i}$ will be described later. 

Without loss of generality, only the counterfactual path of the first unit $Y_{1,t}^0$ after treatment assignment at time $t>T$ is of interest. As in equation (\ref{potential_outcomes}), $Y_{1}^0$ is partially observed through discrete observations in the data up to time $T$ and $Y_{1}^1$ is partially observed after time $T$. All other units are not administered treatment and act as control paths. Let $\mathbf Y^0 = (Y^0_2,\dots,Y^0_n):[t_0,t_m]\rightarrow \mathbb R^{(n-1)\times d}$ be the $(n-1) \times d$ dimensional path that includes all $n-1$ control paths. 

We make the assumption that there exists a continuous function $f: \mathbb R^{d} \rightarrow \mathbb R^{d \times (n-1)}$ such that the counterfactual path of the treated unit $Y_1^0 : [t_0,t_m] \rightarrow \mathbb R^d$ is defined as the solution to the following controlled differential equation (CDE),
\begin{align}
\label{cde_intro}
    Y^0_{1,t} = Y^0_{1,t_0} + \int_{t_0}^{t} f(Y^0_{1,s})\hspace{0.1cm} d\mathbf Y^0_s,\quad t\in(t_0,t_m]
\end{align}
where the integral is a Riemann–Stieltjes integral and "$f(Y^0_{1,s})\hspace{0.1cm} d\mathbf Y^0_s$" is understood as matrix-vector multiplication \cite{kidger2020neural}. (\ref{cde_intro}) is the integral of (\ref{cde_contributions}) where the vector field $f$ in (\ref{cde_contributions}) is taken to act linearly on $d\mathbf Y^0_s$. We say that $Y^0_1$ is controlled or driven by $\mathbf Y^0$, hence the name controlled differential equations.

\textbf{Definition 1} \textit{A continuous-time synthetic control, approximating the counterfactual path $Y_{1,t}^0$ is defined as, 
\begin{align*}
    Y_{1,T}^0 + \int_{T}^{t} f(Y^0_{1,s})\hspace{0.1cm} d\mathbf Y^0_s,\qquad t\in(T,t_m]
\end{align*}
and may be interpreted as a non-linear continuous-time extension to the linear discrete-time synthetic control $\sum^{n}_{i=2} w_i Y_{i,t}^0$ of \cite{abadie2010synthetic}}.

Similarly, the causal effect at time $t>T$ of an intervention administered at time $T$ can be estimated through:
\begin{align*}
    \hat \tau_{1,t} = Y_{1,t}^1 - Y_{1,T}^0 - \int_{T}^{t} f(Y^0_{1,s})\hspace{0.1cm} d\mathbf Y^0_s,\quad t\in(T,t_m], 
\end{align*}
The first term $Y_{1,t}^1$ is observed for $t\in(T,t_m]$ while the integral term is learned by optimizing $f$ such as to approximate the (observed, through irregular samples) $Y_{1,t}^0$ for $t\in(t_0,T)$ before intervention at time $T$.

This strategy may be justified i.e., the treatment effect estimator is unbiased, in the linear case starting with an underlying linear data generating dynamical system, similarly to \cite{abadie2019using}. We show this in the Appendix. In the non-linear case, an estimator of $f$ will typically be biased due to multiple different local minima.

\subsection{Remarks}
We make the following remarks as a comparison to the original synthetic control methodology.
\begin{itemize}[leftmargin=*]
    \item \textbf{Continuous-time.} The proposed formalism explicitly models observed sequences as processes evolving continuously in time. It therefore uses the full information of path observations and time intervals between observations and can be evaluated at any point in time, as illustrated in Figure \ref{overview}.
    \item \textbf{Regularity of dynamics.} The implicit assumption is that the dynamics of the system are regular enough: the dynamics before time $T$ can be extrapolated to the dynamics after time $T$. This is in contrast with the invariance in correlations at \textit{all} times that the vector of weights in equation (\ref{te_eq}) specifies. In weakly coupled dynamical systems we know this assumption to break down with important consequences for the validity of the extrapolation of synthetic controls, as shown in \cite{ding2020dynamical}.
    \item \textbf{Latent state.} In realistic scenarios, it is often the case that observations are a function of an underlying \textit{latent} state whose dynamics follow a differential equation e.g., a country's economy may evolve according to a differential equation although in practice only economics indicators are observed. 
    
    Accordingly, one may define a latent state $g(Y_1^0)=:z_1:[t_0,t_m]\rightarrow \mathbb R^h$ of the counterfactual path $Y_1^0$ as the solution to equation (\ref{cde_intro}), with $h$ the dimension of the latent state. The synthetic control is then the projection of this latent state onto the space of observations, just as indicators are a projection of the latent state of a country's economy. This formalism is described in section \ref{sec_ncsc}.
    \item \textbf{Transparency.} Synthetic controls are desirable also because of their accessible and transparent interpretation. Non-linearities inevitably trade-off some transparency for greater flexibility but we will see that we may regularize the solution space to promote sparsity in the control paths $\mathbf Y^0$ that influence the product $"f(Y^0_{1,s})\hspace{0.1cm} d\mathbf Y^0_s"$. As a result, only few controls drive the counterfactual path of interest, which may be inspected by the user for interpretation.
    \item \textbf{Misaligned observations.} Since only discrete observations are usually available, each path may be approximated in practice with natural cubic splines. One benefit is that this allows for irregular time stamps between units as each path may be interpolated independently.
\end{itemize}

\section{Related work}
\label{sec_related_work}
In light of the remarks above, in this section we review the methodological body of work that has extended the original proposal of Abadie et al. \cite{abadie2010synthetic}, we review the literature on treatment effect estimation with structural models and we review other methods for differential equation modelling.

\textbf{Synthetic controls.} A particular choice for estimating weights depending on data structure and model assumptions is often the distinguishing feature of recent synthetic control methods. For instance, \cite{doudchenko2016balancing} propose to use negative weights and intercept terms, \cite{amjad2018robust,chernozhukov2017exact} propose regularization terms to promote sparsity and robustness, \cite{ding2020dynamical} propose time-varying weights to model changing correlations between variables and \cite{athey2018matrix} interpret counterfactual estimation as a matrix completion problem regularized with matrix norms. However, whilst often improving goodness of fit, matching in discrete time remains difficult with irregularly aligned data i.e., unit observations not aligned in time.

\textbf{Structural models.} The synthetic control literature is often discussed in contrast to structural time series methods, that explicitly fit the trajectory of counterfactual outcomes using lagged outcomes and covariates. These include the g-computation formula, marginal structural models \cite{robins2000marginal,cole2008constructing} and several flexible extensions using neural networks and Gaussian processes including \cite{bica2020estimating,soleimani2017treatment,schulam2017reliable}. One contrast is that structural methods \textit{balance} distributions between treated and control units (relying on regularity of the treated unit trajectory over time to extrapolate counterfactual estimates) while synthetic controls \textit{match} treated units to control units (relying on regularity across units to extrapolate counterfactual estimates). 

There is an important contrast also in the data requirements of these two approaches. For accurate extrapolation structural models require a large number of control paths and many covariates to approximate the underlying causal structure, while synthetic controls require a large number of path observations. Many applications where synthetic controls have proven successful (mostly with $20-40$ control paths, complex dynamics and hidden variables) are inherently not amenable to structural modelling.

\textbf{Differential equation modelling.} Recently, differential equation models for irregular time series data are increasingly commonplace in the machine learning literature. Of note are Neural Ordinary Differential Equations (ODEs) \cite{chen2018neural}, several extensions that modulate the trajectory of interest with incoming data \cite{rubanova2019latent,de2019gru}, and many other proposals that extend the design of vector fields \cite{dupont2019augmented,zhang2020approximation,chen2020learning}, improve optimization performance \cite{li2020scalable} and incorporate processes driven by stochastic noise \cite{tzen2019neural}. 

This description however has not yet found a way into causal inference. This paper proposes the first continuous-time formalism for synthetic control estimation. 


\textbf{Control in reinforcement learning.} The "continuous control" terminology is also found in the reinforcement learning literature to designate physical control tasks with continuous (real valued) action spaces \cite{lillicrap2015continuous}. This is different and not to be confused with the causality definition where the term refers to absence of treatment.

\section{Neural Continuous Synthetic Controls}
\label{sec_ncsc}
In this section, we propose to parameterize $f$ in equation (\ref{cde_intro}) as a neural network with constraints on the sparsity of control path contributions -- an instance of Neural CDEs \cite{kidger2020neural}.  

Neural CDEs are a family of continuous-time models that explicitly define the latent vector field $f_{\theta}$ by a neural network parameterized by $\theta$, and allow for the dynamics to be modulated by the values of an auxiliary path over time. It generalizes the popular Neural ODE formulation of \cite{chen2018neural}, whose dynamics in contrast are fully specified by an initial state, and may be implemented similarly by augmenting the vector field to vary as a function of $\mathbf Y^0_t$ (the set of control paths) as well as $Y_{1,t}$ (the treated path of interest).

Let $\tilde Y_i^0 : [t_0, t_m] \rightarrow \mathbb R^{d+1}$ be the natural cubic spline with knots at $t_0, \dots, t_{m}$ (or more generally at $t_{i,0}, \dots, t_{i,n_i}$) such that $\tilde Y_{i,t_j}^0 = (Y_{i,t_j}^0,t_j)$, for $j=0,\dots,m$. As we observe only a discretization of the underlying process, $\tilde Y_i^0$ is an approximation for which derivatives may be easily computed. 

Let $g_{\eta}:\mathbb R^{d} \rightarrow \mathbb R^{l}$ be a neural network that embeds the observations into a $l$-dimensional latent state $z_{t} := g_{\eta}(Y^0_{1,t})$. Let $f_{\theta}: \mathbb R^{l} \rightarrow \mathbb R^{(n-1) \times l}$ be a neural network parameterizing the latent vector field and let and $h_{\nu}:\mathbb R^{l} \rightarrow \mathbb R^{d}$ be a neural network that defines the observation mechanism projecting the latent state into the observation space to recover an estimate of the counterfactual path $\hat y_{1,t}:=h_{\nu}(z_{t})$. 

We generalize our problem formulation to assume that a \textit{latent} path (instead of the actual observed $Y^0_{1,t}$) can be expressed as the solution to a controlled differential equation of the form,
\begin{align}
\label{cde}
    z_{t} = z_{t_0} + \int_{t_0}^{t} f(z_{s})\hspace{0.1cm} d\mathbf Y^0_s,\quad t\in(t_0,t_m]
\end{align}


\subsection{Interpretability via sparse contributions}
Arguably one of the reasons for the success of synthetic controls is their natural interpretation as a weighted, sparse combination of control paths that can be inspected by the user.

While non-linearities inevitably make the resulting fit more complex and less interpretable, one may enforce sparsity by explicitly including a weighted diagonal matrix that restricts the contribution of control paths. This extension defines the latent counterfactual state as a solution to,
\begin{align}
\label{cde_sparse}
    z_{t} = z_{t_0} + \int_{t_0}^{t} f(z_{s})\hspace{0.1cm} \mathbf W d\mathbf Y^0_s,\quad t\in(t_0,t_m]
\end{align}
where $\mathbf W\in\mathbb R^{(n-1)\times (n-1)}$ is a time-independent diagonal matrix of trainable parameters that are optimized subject to an $l_1$ penalty on its values to encourage sparsity. It is clear then that $Y_{1}^0$ is independent of $Y_{i+1}^0$ if $[\mathbf{ W}]_{ii}=0$, for $i=1,\dots,n-1$. In fact, we have the following proposition, which implies that this constraint precisely identifies the set of CDEs that are independent of $Y_{i}^0$:

\textbf{Proposition 1}. \textit{Consider the class of CDEs $\mathcal C$ defined by (\ref{cde}) that are independent of $Y_{i}^0$ and the class of CDEs $\mathcal C_0$ defined by (\ref{cde_sparse}) such that the $i$-th diagonal entry of $\mathbf W$ is zero. Then $\mathcal C=\mathcal C_0$.}

\textit{Proof.}  Given in the Appendix.

This proposition provides a rigorous way to enforce that a Neural CDE approximation depends only on a few control paths, and implies also that there is no loss of expressivity or approximating power in the estimation of CDEs that are independent of some control paths using this parameterization.

\subsection{Algorithm}

For each estimate of $f_{\theta}$ and $g_{\eta}$ the forward latent trajectory in time that these functions define through (\ref{cde_sparse}) can be computed using any numerical ODE solver:
\begin{align}
    \hat z_{t_1}, \dots , \hat z_{t_k} = 
    \text{ODESolve}(f_{\theta}, \hat z_{t_0}, (t_1,\mathbf Y^0_{t_1}),\dots , (t_k,\mathbf Y^0_{t_k}) )
\end{align}
The goodness of $f_{\theta}$, $g_{\eta}$, and $h_{\nu}$, is then quantified by a loss function $\mathcal L: \mathbb R^d \times \mathbb R^d \rightarrow \mathbb R$ that compares the reconstructed signal $\hat y_{1,t}:=h_{\nu}(z_{t})$ with the observed trajectory values $y_{1,t}$ for $t<T$ (before treatment assignment). A gradient descent algorithm, backpropagating through the ODE solver and the continuous state dynamics, is then used to update all parameters as in \cite{chen2018neural,kidger2020neural}. 

The optimization problem is defined as,
\begin{align}
    \underset{\theta,\eta,\nu, \mathbf W}{\text{arg min}}& \hspace{0.3cm} \mathcal R(f_{\theta},g_{\eta},h_{\nu},\mathbf W), \\ \mathcal R(f_{\theta},g_{\eta},h_{\nu},\mathbf W)&:= \sum_{i:t_i<T} \mathcal L \left(y_{1,t_i}, \hat y_{1,t_i}\right) + \lambda\sum_{j=1}^{n-1} |\mathbf W_{jj}|\nonumber
\end{align}
where $\lambda>0$ is a hyperparameter and $|\cdot|$ denotes the absolute value.

We call this algorithm for estimating the counterfactual trajectory of a treated unit the Neural Continuous Synthetic Control method (\textbf{NC-SC}).

\begin{figure*}[t]
    \centering
    \subfloat[Sample control paths.]{\adjustbox{raise=-3.6pc}{    
        \includegraphics[width=0.25\textwidth]{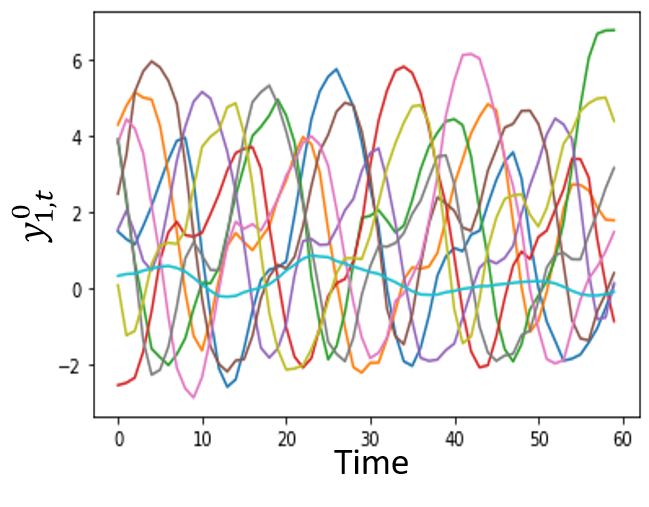}}
    }
    \quad
    \subfloat[Counterfactual estimation performance (lower better) on aligned and mis-aligned observations.]{
        \fontsize{8.5}{9.5}\selectfont
    \centering
    \begin{tabular}{ |p{1.7cm}|C{1.4cm}|C{1.4cm}|C{1.4cm}|C{1.4cm}| }
    
     \cline{2-5}
       \multicolumn{1}{c|}{} & \textbf{Aligned} & \textbf{$30\%$ Dropped} & \textbf{$50\%$ Dropped} & \textbf{$70\%$ Dropped}  \\
     \hline
     SC   & .052 (.003) & .054 (.003)  &  .054 (.003)  & .055 (.003)\\
     \hline
     KMM-SC  & .053 (.003)  & .056 (.004) &  .056 (.005)  & .058 (.005)\\
     \hline
     R-SC  & .051 (.002)  & .053 (.004) &  .056 (.004)  & .055 (.004)\\
     \hline
     MC-SC  & .055 (.003)  & .057 (.003) &  .059 (.002)  & .063 (.003)\\
     \hline
     NC-SC (ours) & .048 (.003)  & .048 (.003) &  .049 (.003)  & .049 (.003)\\
     \hline
    \end{tabular}
    }
    \caption{Experiments on Lorenz's model.}
    \label{lorenz_perf}
\end{figure*}

\section{Experiments}
In this section we experiment with synthetic data from Lorenz's chaotic dynamical system and discuss 2 studies that have received attention in the public policy literature.

\textbf{Evaluation metric.} In all experiments, we report mean and standard deviations of the \textit{control} estimation error \begin{align}
    \frac{1}{|\mathcal T|}\sum_{t\in\mathcal T}||y^0_{1,t} - \hat y^0_{1,t}||_2^2
\end{align}
over 10 model runs, where for real data $\mathcal T = \{t_i:t_i<T, i=2,\dots,n\}$ is the pre-intervention observation times (where untreated data is observed). For synthetic data we use all observation times $\mathcal T = \{t_1, \dots, t_m\}$ for evaluation (as we are free to generate any amount of data).

\textbf{Evaluation methods.} Comparisons are made with the original approach of \cite{abadie2010synthetic} with weights constrained to be non-negative and summing to 1 (\textbf{SC}), with an extension that instead matches pre-treatment outcomes in a reproducing kernel Hilbert space using an instance of kernel mean matching \cite{gretton2009covariate} (\textbf{KMM-SC}), with robust synthetic controls (\textbf{R-SC}) that use penalized weight estimation as in \cite{doudchenko2016balancing}, and with the matrix completion (\textbf{MC-SC}) approach of \cite{athey2018matrix}. 

For experiments involving misaligned data, all methods except NC-SC require some form of prior interpolation and evaluation on a regular grid of time points. Here we use cubic spline interpolations with knots at observation times and smoothing parameter chosen visually for good fit.

Precise experimental details including neural network architectures, optimization software, implementation details and data sources may be found in the Appendix.

\subsection{Lorenz's chaotic model} 
We begin by demonstrating the efficacy of NC-SC on \textbf{irregularly aligned time series} from Lorenz's model for chaotic dynamical systems \cite{lorenz1996predictability}.

The dynamics in a $d$-dimensional Lorenz model are,
\begin{align*}
    \frac{d}{dt}x_i(t) =  \left (x_{i+1}(t) - x_{i-2}(t) \right)\cdot x_{i-1}(t) - x_i(t) + F,
\end{align*}
for $i=1,\dots,d$, where $x_{-1}(t) := x_{d-1}(t)$, $x_0(t) := x_p(t)$, $x_{d+1}(t) := x_1(t)$ and $F$ is a treatment variable that has the effect of changing the level of non-linearity and chaos in the series. We take $F=5$ (mild chaotic behaviour) as the baseline control behaviour and $F=10$ to define the dynamics of the treated regime. The initial state of each variable is sampled from a standard Gaussian distribution and $d$ is set to $10$.

\textbf{Experiment design.} For simplicity, only the counterfactual trajectory of the \textit{first} dimension of the system is of interest, $y^0_{1,t}:=x_1(t)$, while control trajectories are each similarly defined but with \textit{different} random initializations of Lorenz's model. That is, $y^0_{2,t}:= x_1(t)$ with some random initialization, $y^0_{3,t} := x_1(t)$ with some different random initialization and so on (this is equivalent to having units with different features in static models). We set the number of control paths to 20. 

The problem is to construct a synthetic control for the treated unit had $F=5$ for all $t$, given that we observed 200 time observations ($t<200$) with $F=5$ before treatment assignment at time $T=200$. 

Sequences of observations from these paths are observed in two configurations. 
\begin{enumerate}[leftmargin=*, itemsep=0pt]
    \item \textit{Regularly aligned} with a fixed grid of observation times.
    \item \textit{Irregularly aligned} by removing randomly $30\%$, $50\%$ and $70\%$ of the aligned data, independently for each unit.
\end{enumerate}

\textbf{Results.} Performance is computed on a held-out segment of the data (extrapolating the counterfactual path of the treated unit over $t\in(200,400)$). Performance results, as well as an illustration of control paths is given in Figure \ref{lorenz_perf}. Continuous-time synthetic controls outperform every other model considered and furthermore have relatively stable performance with irregular data while other methods exhibit a decrease in performance which we hypothesize is due to worsening imputation performance.

\begin{table}[H]
\fontsize{9.5}{10.5}\selectfont
\centering
\begin{tabular}{ |p{2cm}|C{1.8cm}|C{1.8cm}| }

 \cline{2-3}
   \multicolumn{1}{c|}{} & \textbf{Smoking} & \textbf{Eurozone}   \\
 \hline
 SC   & .0248 (.00) & .0339 (.00)  \\
 \hline
 KMM-SC  & .0221 (.00)  & .0321 (.00) \\
 \hline
 R-SC  & .0002 (.00)  & .0230 (.00) \\
 \hline
 MC-SC  & .0005 (.00)  & .0299 (.00) \\
 \hline
 NC-SC (ours) & .0001 (.00)  & .0003 (.00)\\
 \hline
\end{tabular}
\caption{Counterfactual estimation performance (lower better).}
\label{other_perf}
\end{table}

\subsection{The Eurozone and current account deficits}
Next, we consider an experiment that further highlights the need for \textbf{non-linear combinations} of control paths to accurately approximate the control path of the treated unit.

\textbf{Experiment design.} The problem is to evaluate the impact of Eurozone membership on the path of current account deficits, thought to have considerably aggravated the recovery after the 2007-2008 financial crisis.

By the end of 2009, Europe was at the beginning of a multiyear sovereign debt crisis, in which several Eurozone members were unable to repay their government debt or to bail out over-indebted banks. The eurozone crisis is thought to have been caused in part by a sudden stop of foreign capital investments into countries that had substantial deficits, fueled by low borrowing costs as a consequence, arguably, of Eurozone membership \cite{frieden2017understanding}. One may ask then whether there is any evidence for this claim, whether or not a country's current account deficits would have been different had it not join the Eurozone.   

\begin{figure*}[t]
    \centering
    \subfloat[Baseline comparisons.]{\centering
    \includegraphics[width=0.45\textwidth]{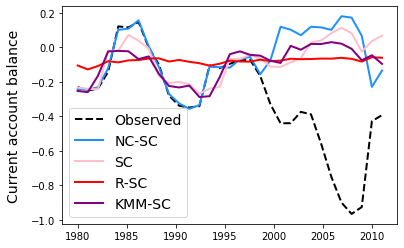}
    }
    \qquad
    \subfloat[Contrast with the trajectory of the most influential control countries.]{
        \centering
    \includegraphics[width=0.45\textwidth]{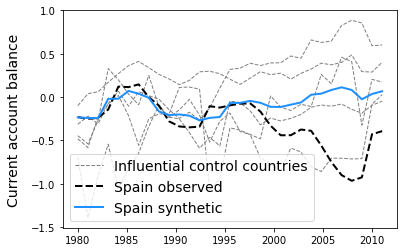}
    }
    \caption{Counterfactual current account predictions for Spain $\hat Y_{1,t}^0$ over time. The euro was made effective in 1998. Influential countries are Chile, Hungary, Japan, Mexico and Sweden.}
    \label{emu}
\end{figure*}

We focus on one country, Spain. The data consists of yearly current account figures from 1980 to 2010 for Spain as well as 15 other countries outside the Eurozone, as collected by David Hope in \cite{hope2016estimating}. The pre-treatment period ranges from 1980 to 1998 when the Eurozone was made effective (as illustrated in Figure \ref{emu}). 

\textbf{Results.} Performance results are given in Table \ref{other_perf}. These demonstrate that NC-SC can substantially improve performance in this case, as also illustrated in the model fit to the observed control trajectory in Figure \ref{emu}. 

Figure \ref{emu} shows which other countries were influential in determining the synthetic control for Spain, inferred by inspecting the non-zero entries of $\mathbf W$. NC-SC uses combinations of current account balance figures from Chile, Hungary, Japan, Mexico and Sweden, in contrast to those of Great Britain, Israel, Mexico and Sweden used by the original synthetic control method \cite{abadie2010synthetic}. Interestingly, as a result, the projection given by NC-SC gives a slightly different interpretation estimating a positive current account balance had Spain not adopted the Euro in contrast to a zero current account balance given by \cite{abadie2010synthetic}. 

\subsection{Smoking control in California}

Next, we consider one of the most popular benchmarks for synthetic control estimation, namely the evaluation of the effect of the influential 1988 anti-smoking legislation in California on cigarette sales. 

\textbf{Experiment design.} At that time, California lead a wave of anti-smoking legislation, known as Proposition 99, that served as a model for policy interventions in other states later on and arguably reduced the prevalence of smoking. The problem is to assess its effect in comparison to California's cigarette sales had the legislation not been passed.

We follow the experiment by \cite{abadie2010synthetic} and use annual state-level panel data for the period 1970-2000, giving us 19 years of pre-intervention cigarette sales data. 

\textbf{Results.} Performance comparisons are given in Table \ref{other_perf} and the corresponding fit and treatment effect (as the difference between the counterfactual estimation and observed trajectory) is illustrated in Figure \ref{smoking}. Continuous-time synthetic controls, as well as baseline methods match almost exactly the pre-treatment trajectory of the treated unit and all counterfactual projections point towards an important treatment effect. The California anti-smoking legislation was responsible for part of the lowering of cigarette sales.

We show in addition the contribution of each state to the synthetic control in Figure \ref{smoking}, inferred by inspecting the non-zero entries of $\mathbf W$. In this case there is little contrast with existing baselines; most methods use Nevada, Utah, Montana, Colorado and Connecticut as the most influential states for the construction of synthetic controls which serves to confirm the estimates of NC-SC.



\section{Discussion}

We conclude with some additional remarks and clarifications that may be of practical importance. 

\begin{itemize}[leftmargin=*,topsep=-1pt]
\item \textbf{On the role of covariates.} Auxiliary covariates have not played a role in the development of continuous-time synthetic controls. While \cite{abadie2010synthetic} demonstrate the treatment effect to be asymptotically unbiased under a perfect match on both pre-treatment outcomes and relevant covariates (among other conditions) this is not strictly necessary as long as sufficient pre-treatment outcomes are observed (see Theorem 1 \cite{botosaru2019role}). The intuition behind this result is that it would not be possible to match on a large number of pre-treatment outcomes without matching on both observed and unobserved relevant covariates. 

However, if desired, matching on time-variant or time-invariant covariates to define continuous-time synthetic controls with our formalism is possible and straightforward by using a data-dependent lasso regularization scheme on the matrix $\mathbf W$. Instead of penalizing all entries of $\mathbf W$ equally, one may adopt a relevance weighting approach $\sum_{i=1}^{n-1}\frac{1}{\hat p_i}|\mathbf W_{ii}|$, where $\hat p_i > 0$ measures the relevance of the $i$-th control covariates towards matching the covariates of the treated unit (higher values of $\hat p_i$ corresponding to more relevant control paths in this case). 

For example, with $n$ units of $d$-dimensional static covariates $\mathbf X \in \mathbb R^{n\times d}$ observed along the outcome paths of interest, $\hat p = (\hat p_2,\dots,\hat p_n)$ may be defined by the linear projection of $\mathbf X_{1}$ onto $\mathbf X_{2:n}$, $\hat p = (\mathbf X_{2:n}^T \mathbf X_{2:n})^{-1}\mathbf X_{2:n}^T\mathbf X_{1}$. 


\begin{figure*}[t]
    \centering
    \subfloat[Baseline comparisons.]{\centering
    \includegraphics[width=0.45\textwidth]{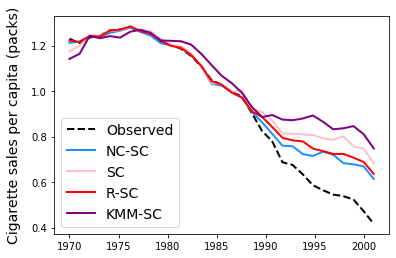}
    }
    \qquad
    \subfloat[Contrast with the trajectory of the most influential control states.]{
        \centering
    \includegraphics[width=0.45\textwidth]{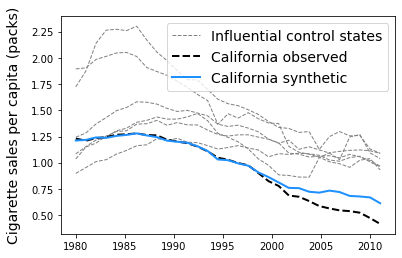}
    }
    \caption{Counterfactual cigarette sales predictions for California $\hat Y_{1,t}^0$ over time. Anti-smoking legislation was introduced in 1988.}
    \label{smoking}
\end{figure*}

\item \textbf{Control population.} The similarity between the control population and the treated unit in the pre-treatment period and the posterior counterfactual trajectory underlies the validity of synthetic controls. The choice of control population is therefore important and there are two important requirements that must be met.  

A first requirement is that control units not be affected by the intervention or treatment of interest so that they faithfully describe the counterfactual trajectory of the treated unit. This assumption has to be justified and is not necessarily always plausible. For instance one may argue that public policy interventions have spillover effects e.g., cigarette sales of neighbouring control states to California being affected by anti-smoking legislation or the economy of control countries with strong commercial ties to Eurozone members influenced by the monetary union, in which case counterfactual estimates will be biased. In these particular two examples, this assumption however has been carefully justified \cite{abadie2010synthetic, hope2016estimating}.  

A second requirement is that units, after the intervention, not be subject to large idiosyncratic shocks that would not have affected the treated unit in the absence of treatment as such control units would not remain representative of the counterfactual trajectory. This assumption relates to the regularity of the correlations between control and treated units over time, which holds if a common underlying causal model for the data can plausibly be assumed.

\item \textbf{Data requirements.} The credibility of synthetic controls hinges on the accuracy of pre-treatment control path approximation. Therefore a sizeable number of pre-treatment observations should be available for valid extrapolations. This is perhaps in contrast with structural models that require observation of all covariates and a larger number of control paths for accurate extrapolation. 

We have omitted an explicit comparison with structural models because of this practical difference. If not all variables in the data generating mechanism are observed, as in our experiments (for instance, we do not have any information on the driving forces that determine the current account balance in the Eurozone experiment), it is not plausible to fit a structural equation model. 

\item \textbf{Uncertainty estimation}. As presented here, continuous synthetic controls do not explicitly quantify uncertainty in counterfactual estimation. Such extensions are feasible given that stochastic differential equations (SDEs) can be expressed as a controlled differential equation driven by a stochastic process, and given existing work on backpropagating through SDE solvers that may be used with a neural vector field in analogy to Neural ODEs \cite{kong2020sde,liu2019neural}.

\end{itemize}

\section{Conclusion}
This paper has demonstrated how synthetic control estimation may be extended to continuous-time using the mathematics of controlled differential equations. 

Our proposal for counterfactual estimation, called Neural Continuous Synthetic Controls, models explicitly the latent paths of the observed time series defining a synthetic control as a combination of paths rather than as a combination of discrete observations. Neural Continuous Synthetic Controls are conceptually natural for modelling processes continuously unfold over time and accommodate for irregularly aligned data and more complex dynamics than previously analysed.

\bibliography{bibliography}
\bibliographystyle{plain}

\newpage
\appendix
{\Large \textbf{Appendix}}
\\\\
This appendix provides additional material accompanying the main body of the paper: "Synthetic Controls in Continuous time". It is outlined as follows:

\begin{itemize}
    \item Section \ref{sec_review} presents proofs of the theoretical results presented in the main body of this paper.
    \begin{itemize}
        \item Section \ref{sec_bias} provides an argument for the unbiasedness of treatment effects in continuous-time for a linear model, in analogy to \cite{abadie2010synthetic}.
        \item Section \ref{sec_proof} proves Proposition 1.
    \end{itemize}
    \item Section \ref{sec_additional_experiments} discusses the reliability of inference and computational complexity.
    \begin{itemize}
        \item Section \ref{sec_w_consistency} discusses the consistency of the recovered matrix $\mathbf W$ in different runs of the algorithm.
        \item Section \ref{sec_fit_consistency} discusses the consistency of NC-SC's fit in different runs of the algorithm.
        \item Section \ref{sec_complexity} discusses computational complexity.
    \end{itemize}
    \item Section \ref{sec_implementation} gives details on software and algorithm implementation.
    \item Section \ref{sec_experiments} gives further details on the design of experiments and gives pointers to the publicly available data and simulation environments.
\end{itemize}

\section{Theoretical results}
\label{sec_review}

\subsection{Unbiased treatment effects in linear dynamical systems}
\label{sec_bias}
Synthetic controls and their use for treatment effect estimation is typically justified in the discrete-time setting by an underlying linear data generating mechanism involving observed, unobserved variables, and independent noise terms \cite{abadie2010synthetic}. Extending this analysis to non-linear models is difficult and to our knowledge has not be done even in the discrete-time case. 

We may provide some theoretical justification for the validity of NC-SC however by considering the continuous-time analog to the autoregressive linear model considered in \cite{abadie2010synthetic},
\begin{align}
\label{abadie}
    y_{i,t_{s+1}} = \alpha_{t}y_{i,t_{s}} + z_{i,t_{s}}, \quad \text{for} \quad i=1,\dots,n \quad \text{and} \quad s=1,\dots,m,
\end{align}
where $z_{i,t}$ is not modelled and may be correlated across units and time to account due to unobserved confounding but is assumed to have mean zero conditional on $\mathcal F_t = \{y_{i,t}\}_{i=1,\dots,n, t=t_1,\dots,t_m}$.

A continuous-time analog is given by,
\begin{align}
    \frac{dy_i(t)}{dt} = \alpha(t) y_i(t) + z_i(t), \qquad y_i(t_0) = y_{i,0}
\end{align}
for $t \in (t_0,t_m)$. All parameters and variables now vary continuously in time and the assumption, in parallel to equation (\ref{abadie}) is that confounders $z_{i}(t)$ have mean zero conditional on $\mathcal F_t = \{y_{i}(s): s<t\}_{i=1,\dots,n}$.

Following the steps taken in \cite{abadie2010synthetic} we may consider an infinitessimal change in time $\Delta t>0$ and write,
\begin{align}
\label{infinitessimal}
    \frac{dy_i(t+\Delta t)}{dt} &= \alpha(t+\Delta t) y_i(t+\Delta t) +  z_i(t+\Delta t)
    \\
    &= \alpha(t+\Delta t) \left(y_i(t) + \Delta t \frac{dy_i(t)}{dt}\right) + z_i(t+\Delta t)
\end{align}
Assume now that there exists weights $w_2^{\star}, \dots, w_n^{\star}$ such that for all times $t$ before the intervention time $T$ i.e., those times where we do have data and can compare the derivatives of the path of interest with control paths,
\begin{align}
\label{weights}
    \frac{dy_1(t)}{dt} = \sum_{i=2}^n w_i^{\star}\frac{dy_i(t)}{dt}, \qquad y_{1,0} = \sum_{i=2}^n w_i^{\star}y_{i,0}, \qquad t<T
\end{align}
It hols then that for $T = \Delta t + t$ for $\Delta t >0$ and $t<T$,
\begin{align}
    \frac{dy_1(T)}{dt} - \sum_{i=2}^n w_i^{\star}\frac{dy_i(T)}{dt} &=  \alpha(t+\Delta t) \left( y_1(t) - \sum_{i=2}^n w_i^{\star}y_i(t) + \Delta t \left( \frac{dy_1(t)}{dt} - \sum_{i=2}^n w_i^{\star}\frac{dy_i(t)}{dt}\right)\right) \nonumber\\
    &+ z_1(T) - \sum_{i=2}^n z_i^{\star}y_i(T)
\end{align}
Then, using the relationship in (\ref{weights}), working recursively on the difference $y_1(t) - \sum_{i=2}^n w_i^{\star}y_i(t)$, and the fact that $z_1(T) - \sum_{i=2}^n z_i^{\star}y_i(T)$ has mean zero conditional on the filtration $\mathcal F_t$, the above quantity has mean zero.

We may then extend this result to $t>T$ by using equation (\ref{infinitessimal}).

\cite{abadie2010synthetic} also analyze factor models but since they do not incorporate time-varying confounders, observed or unobserved, we do not replicate their analysis here.

\subsection{Proof of Proposition 1}
\label{sec_proof}
We restate Proposition 1 for completeness, and redefine the two CDe models of interest.

We consider models of the form,
\begin{align}
\label{cde2}
    z_{t} = z_{t_0} + \int_{t_0}^{t} f(z_{s})\hspace{0.1cm} d\mathbf Y^0_s,\quad t\in(t_0,t_m]
\end{align}
and models of the form,
\begin{align}
\label{cde_sparse2}
    z_{t} = z_{t_0} + \int_{t_0}^{t} f(z_{s})\hspace{0.1cm} \mathbf W d\mathbf Y^0_s,\quad t\in(t_0,t_m]
\end{align}
with the same notation as in the main body of this paper.

\textbf{Proposition 1}. \textit{Consider the class of CDEs $\mathcal C$ defined by (\ref{cde2}) that are independent of $Y_{i}^0$ and the class of CDEs $\mathcal C_0$ defined by (\ref{cde_sparse2}) such that the $i$-th diagonal entry of $\mathbf W$ is zero. Then $\mathcal C=\mathcal C_0$.}

\textit{Proof.}  It is clear that the class of Neural CDEs $\mathcal C_0$ is contained in $\mathcal C$ as control paths interact with the vector field of the counterfactual path only if the corresponding entry in $\mathbf W$ is non-zero. $\mathcal C_0 \subset \mathcal C$.

For the converse consider a CDE in $\mathcal C$ defined by (\ref{cde2}) independent of the $k$-th control path and consider a set of control paths $\tilde{\mathbf Y}^0_s$ and $\mathbf Y^0_s$ such that $\tilde{\mathbf Y}^0_s = \mathbf Y^0_s$ except for the $k$-th control path of $\tilde{\mathbf Y}^0_s$ which is set to the zero function $\tilde Y^0_{k}:[t_0,t_m]\rightarrow 0$, for all $s\in[t_0,t_m]$. Because of independence, the CDEs with control paths $\tilde{\mathbf Y}^0_s$ and $\mathbf Y^0_s$ are equal.

Define the diagonal matrix $\tilde{\mathbf W}$ and $\mathbf W$ such that their diagonal entries agree except on their $k$-th diagonal entry of $\tilde{\mathbf W}$ where $[\tilde{\mathbf W}]_{kk}=0$. Then we can see that $\tilde{\mathbf W}d\mathbf Y^0_s = \mathbf W d\tilde{\mathbf Y}^0_s$, so that the two CDEs given by (\ref{cde_sparse2}) define the same functions. A CDE (\ref{cde2}) independent of one of its arguments may always be reconstructed with a corresponding CDE (\ref{cde_sparse2}). This implies that the class of Neural CDEs independent of the $k$-th control path is contained in $\mathcal C_0$. $\mathcal C \subset \mathcal C_0$ \qed

\section{Additional experiments}
\label{sec_additional_experiments}

\subsection{Consistency of $\mathbf W$}
\label{sec_w_consistency}
Different runs of the algorithm may converge to different local minima thus potentially changing the interpretation of the matrix $\mathbf W$ of control path contributions. We did observe slight differences in the entries of the estimated matrix $\mathbf W$ although its interpretation: which entries were estimated to zero and non-zero,  remained consistent. 

We tested this feature with the current account deficit data. We ran NC-SC 10 times with different initializations and report the variance of estimated weights for each country in Table \ref{w_consistency}. For almost all entries the weight variance is consistently low and no run gives a qualitatively different interpretation (for instance one run estimating $\mathbf w_i$ to zero while another run estimating it to a value different than zero). 

While this is not an exhaustive test of this behaviour it provides some evidence that $\mathbf W$ may be used consistently to recover the control paths most influential in counterfactual estimation.

\begin{table}[H]
\fontsize{8.5}{9.5}\selectfont
\centering
\begin{tabular}{ |C{0.8cm}|C{0.8cm}|C{0.8cm}|C{0.8cm}|C{0.8cm}|C{0.8cm}|C{0.8cm}|C{0.8cm}|C{0.8cm}|C{0.8cm}|C{0.8cm}|C{0.8cm}|C{0.8cm}|C{0.8cm}|C{0.8cm}| }
 \hline
  CHL & DNK & HUN & ISR & JPN & MEX & NZL & POL & SWE & USA & Others \\
 \hline
 .127 (.02) & .063 (.01) & .076 (.01) & .068 (.02) & .123 (.04) & .105 (.03) & .051 (.02) & .096 (.02) & .089 (.01) & .088 (.02) & .000 (.00)  \\
 \hline
\end{tabular}
\caption{Weight estimation over 10 runs of NC-SC. Others are: Canada, Turkey, Great Britain, Australia and Korea.}
\label{w_consistency}
\end{table}

\subsection{Consistency of NC-SC fit}
\label{sec_fit_consistency}
Just as different runs of NC-SC may produce different estimated matrices $\mathbf W$, may result in different counterfactual fits. Using the same experiment as above we show that NC-SC's fit is remarkably consistent across different initializations.

\begin{figure*}[t]
    \centering
    \subfloat[NC-SC fit over 5 runs with different initializations on the current account deficit data for Spain.]{   
        \includegraphics[width=0.3\textwidth]{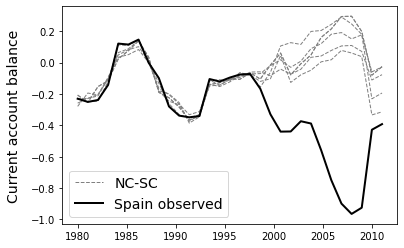}
    }
    \hfill
    \subfloat[Run time of NC-SC as a function of the number of control paths.]{
        \includegraphics[width=0.3\textwidth]{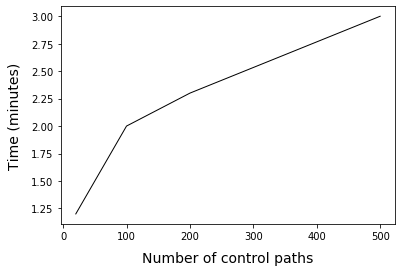}
    }
    \hfill
    \subfloat[Run time of NC-SC as a function of the number of pre-treatment observations.]{   
        \includegraphics[width=0.3\textwidth]{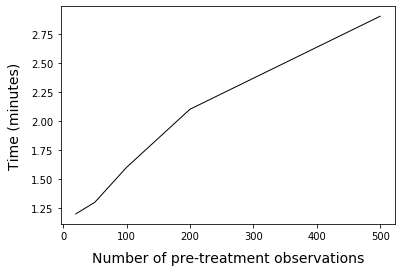}
    }
    \caption{Analysis of fit consistency and computational complexity.}
    \label{fig_consistency}
\end{figure*}

Figure \ref{fig_consistency} shows the fit of NC-SC in 5 runs with different initialization parameters. We used the current account deficit data for this experiment.

\subsection{Computational complexity}
\label{sec_complexity}
Run times of NC-SC as a function of the number of pre-treatment observations and as a function of the number of control paths is given in Figure \ref{fig_consistency}. We used the Lorenz model for this experiment.

\section{Details on algorithm implementation}
\label{sec_implementation}
This section gives details on the implementation of our algorithm as well as implementation software of baseline methods.

\subsection{Neural Continuous Synthetic Controls}
For completeness, this section reviews our modelling choices for Neural CDEs and alternatives, following the analysis of \cite{kidger2020neural}.

NC-SC is defined using the Neural CDE,
\begin{align}
\label{cde_sparse2}
    z_{t} = z_{t_0} + \int_{t_0}^{t} f(z_{s})\hspace{0.1cm}\mathbf W\hspace{0.1cm} d\mathbf Y^0_s,\quad t\in(t_0,t_m]
\end{align}
with the notation used in the main body of this paper.

\begin{itemize}
    \item \textbf{Path interpolation.} We followed \cite{kidger2020neural} in approximating the underlying paths $\mathbf Y^0_s$ using cubic spline interpolations with knots at the observation times. The minimum requirement for evaluating the CDE in equation (\ref{cde_sparse2}) is therefore that $\mathbf Y^0_s$ be at least continuous and piecewise differentiable.

    However, training with the adjoint backpropagation method requires derivatives of a functional of the CDE with respect to time, an thus second derivatives of the path approximations $\mathbf Y^0_s$. For this to be done consistently, the choice of cubic splines essentially gives the minimum smoothness requirement to paths approximations. 
    
    \item \textbf{Other vector field choices.} If one wanted to incoporate the influence of auxiliary paths to modulate the trajectory of $z$, different choices for the vector field $"f(z_{s})\hspace{0.1cm} d\mathbf Y^0_s"$ could have been made. For instance, \cite{de2019gru} developed a time series method defining the vector field as $g(z_s,Y^0_s)$ for some function $g$ to be learned. This is perhaps a more natural choice that considers a function of paths $\mathbf Y^0_s$ explicitly and allows for non-linearities in the interaction between $z$ and $\mathbf Y^0_s$. \cite{kidger2020neural} showed, however, that the vector field $"f(z_{s})\hspace{0.1cm} d\mathbf Y^0_s"$ is strictly more general, subsuming models of the form $g(z_s,Y^0_s)$.

    \item \textbf{Architecture.} The integrand $f$ was taken to be a feed-forward neural network with a two hidden layers of size 10 and elu activation functions after each layer except from the output layer. The dimensionality $l$ of $z$ the hidden state was taken to be $5$ for all experiments. The activation function was chosen to be the \texttt{elu} function, although the \texttt{relu} performed similarly but was less stable in optimization with a larger variance across different runs of the algorithm.
    
    We did not explicitly tune hyperparameters (hidden layers, activation function, etc.) for performance. Further tuning could be done by cross-validation on the observed pre-treatment trajectory of the unit of interest if enough observations are given. 
    
    \item \textbf{Optimization.} In each case we used the Adam optimiser as implemented by PyTorch. Starting learning rates varied between experiments (with values between 0.001 and 0.01) before being reduced by half if metrics failed to improve for a certain number of epochs.
    
    The strength of $\mathbf W$ regularization $\lambda$ is chosen in a range $\{0.001,0.01,0.1,1\}$ for best performance on a validation set.
    
    The ODE solver used to extrapolate the hidden state was taken to be the fourth-order Runge-Kutta with 3/8 rule solver, as implemented by passing \texttt{method='rk4'} to the \texttt{odeint\_adjoint} function of the \texttt{torchdiffeq} \cite{kidger2020neural,chen2018neural} package, used also for adjoint back-propagation training. The step size was taken to equal the minimum time difference between any two adjacent observations.
\end{itemize}

\subsection{Original Synthetic Controls}
We implement the original synthetic control method \cite{abadie2010synthetic} with the \texttt{cvxpy} python package for constrained convex optimization. This allows us to enforce weights to be non-negative and sum to one in few straightforward lines of code.

\subsection{Robust Synthetic Controls}
We implement a penalized version of the original synthetic control method with an elastic net penalty and hyperparameters chosen by cross-validation with the \texttt{sklearn} python library and predefined optimization function \texttt{ElasticNetCV}.

\subsection{Synthetic Controls with Kernel Mean Matching}
We implement a modification of the original synthetic control method, instead matching weights in a reproducing kernel Hilbert space with Gaussian kernel and bandwidth parameter taken to be the median distance between any pair of observations as suggested by the authors \cite{gretton2009covariate}. Our implementation code follows exactly the procedure given in the constrained optimization program defined in section 1.1.3 of \cite{gretton2009covariate}.

The algorithm itself was implemented in python with a constrained quadratic program \texttt{cvxpy}, similarly to the original synthetic control method.

\subsection{Matrix Completion}
We implement the matrix completion method with nuclear norm penalization of \cite{athey2018matrix} with the \texttt{matrix completion} package in python. Its implementation is straightforward with this package taking only a single line of code to define and fit the model.

\section{Experiment details}
\label{sec_experiments}

\subsection{Smoking and Eurozone experiments}

The data from both the smoking study in California and current account deficits in Spain is annualized. For a period of 30 years we thus have 30 observations on each trajectory (e.g. cigarette sales in California, Nevada, etc., and current account deficits in Spain, Switzerland, etc.) and in particular in both studies we have 19 years / observations of pre-intervention data that we may use for quantitative performance evaluations. 

This is rarely sufficient for reliable comparisons between algorithms. Our approach to correct for this problem is to augment the data by interpolating trajectories with cubic splines with knots at the observation times. The smoothness hyperparameter is chosen such that the interpolation accurately reflects the original data trajectory. With this interpolation, we evaluate the curve for each path at 300 regular points spanning the 19 years of pre-intervention data. 

In our experiment we use the first 200 observations for training and we use the last 100 observations for testing, on which we report performance.

\end{document}